\def\affiliationcopy{}
\documentclass{article}

\usepackage{iclr2027_conference,times}

\usepackage{amsmath,amssymb,amsthm}
\usepackage{booktabs}
\usepackage{multirow}
\usepackage{array}
\usepackage{graphicx}
\usepackage{xcolor}
\usepackage{colortbl}
\definecolor{tencentblue}{HTML}{0052D9}
\definecolor{tencentwash}{HTML}{E7F1FC}
\definecolor{tencentrow}{HTML}{C5D8F7}
\definecolor{mabhead}{HTML}{F3F3F3}
\definecolor{mablong}{HTML}{E6E2D8}
\definecolor{mabrag}{HTML}{F2E6D4}
\definecolor{mabagent}{HTML}{EADFE6}
\providecommand{\mabb}[1]{\textbf{#1}}
\providecommand{\mabu}[1]{\underline{#1}}
\providecommand{\thiswork}[1]{\textbf{\textcolor{tencentblue}{#1}}}
\providecommand{\mabsetup}{%
  \setlength{\tabcolsep}{4.2pt}%
  \renewcommand{\arraystretch}{1.12}%
  \setlength{\aboverulesep}{0pt}%
  \setlength{\belowrulesep}{0pt}}
\providecommand{\mabband}[3]{%
  \cellcolor{#2}\rule[-2.6pt]{0pt}{11pt}%
  & \multicolumn{#1}{c}{\cellcolor{#2}{\itshape #3}}}

\usepackage{tikz}
\usetikzlibrary{arrows.meta,positioning,fit,backgrounds}
\usepackage{float}
\usepackage{algorithm}
\usepackage{algorithmic}
\usepackage{hyperref}
\usepackage{url}
\hypersetup{hidelinks}
\pdfpageattr{/Group << /S /Transparency /I true /CS /DeviceRGB >>}

\newcommand{\method}{\textsc{CMC}}

\title{Contract Memory Compiler:\\
Resolve, Then Traverse}

\ifdefined\affiliationcopy
  \iclrfinaltrue
  \makeatletter
  \def\@maketitle{\vbox{\hsize\textwidth
    {\LARGE\sc \@title\par}
    \vskip 0.06in
    {\centering\small\bf
      Zhi Song$^{1,2}$\quad XiMing Xing$^{2}$\quad Chunhan Li$^{2}$\quad Weian Mao$^{3}$\\[0.12em]
      Zhenchao Tang$^{2}$\quad Hanbo Huang$^{2}$\quad Fan Xu$^{2}$\quad Jiale Zhou$^{2}$\\[0.12em]
      Jiahui Guan$^{2}$\quad Zejian Ding$^{1}$\quad Chen Ma$^{1}$\quad Lusheng Wang$^{1}$\par}
    \vskip 0.28em
    {\centering\footnotesize
      $^{1}$Department of Computer Science, City University of Hong Kong, Hong Kong SAR\\[0.08em]
      $^{2}$Tencent, China\\[0.08em]
      $^{3}$Department of Electrical Engineering and Computer Science, Massachusetts Institute of Technology, USA\par}
    \vskip 0.08in}}
  \makeatother
  \AtBeginDocument{%
    \fancyhf{}%
    \AddToShipoutPictureFG*{%
      \AtTextUpperLeft{%
        \put(0,16){%
          \raisebox{0.41cm}{\scalebox{1}[-1]{\includegraphics[height=0.41cm]{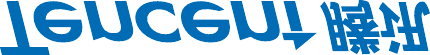}}}%
          \hspace{0.7cm}%
          \raisebox{0.94cm}{\scalebox{1}[-1]{\includegraphics[height=0.94cm]{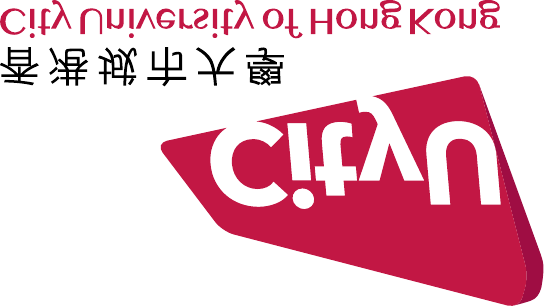}}}%
        }%
      }%
    }%
  }
  \author{Zhi Song, XiMing Xing, Chunhan Li, Weian Mao, Zhenchao Tang, Hanbo Huang, Fan Xu, Jiale Zhou, Jiahui Guan, Zejian Ding, Chen Ma, and Lusheng Wang}
\else
  \author{Anonymous Authors}
\fi

\begin{document}
\maketitle
\suppressfloats[t]

\begin{abstract}
External memory lets language-model agents answer questions about histories
too long for the answer model's context window. Updates create a harder problem than
retrieving a recent fact: changing one relation can redirect a multi-hop
question to records about an entity absent from the question. We study this
\emph{update-dependent evidence selection} problem and introduce the
Contract Memory Compiler (\method{}). Before seeing a question, \method{}
uses a language model to identify relations in the history and record where
each one was stated. It applies later updates to determine the current relations,
follows them from entities named in the question, and passes the corresponding
original records to the answer model in one call.
Thus the current state determines which evidence is read, rather than merely
refreshing values in a previously selected context. To the best of our
knowledge, \method{} achieves state-of-the-art multi-hop accuracy on
FactConsolidation, reaching 78.25\% overall and 61.0\% at 262K.
With the extracted relations and answer model held fixed, selecting evidence
before resolving updates reduces multi-hop accuracy to 21.50\%.
We also introduce MQuAKE-MemStream, a derived dataset of ordered memory
streams built from MQuAKE-Remastered counterfactual cases.
\end{abstract}

\section{Introduction}

A language model agent may interact with the same user for months. It learns
a preference in one session and a decision in another. The user may later
revise that decision. The model cannot keep the entire history in its active
context, and long inputs can still hide evidence in the middle
\citep{liu2024lostmiddle}. External memory stores information across sessions. It supplies a
small set of records for each response. The agent must therefore record useful
information, track changes, and find the right records later. Benchmarks for
persistent memory test reasoning across sessions, knowledge updates, selective
forgetting, and retrieval
\citep{wu2024longmemeval,hu2026memoryagentbench}.

Memory systems address different parts of this process.
Compression and synthesis turn interactions into compact, searchable records
\citep{liu2026simplemem}. Linked notes and graph memories preserve connections
among events and entities \citep{wu2026gam,ong2025theanine}. Temporal memory
systems retain past versions of facts and identify the current version
\citep{banerjee2026apexmem,cui2026memtxn}. Some agents choose storage, update,
and retrieval operations as part of their workflow \citep{yu2026agemem}.
These designs determine what persists and how it can be accessed. At answer
time, the model sees only the records placed in its current input. We study
how to choose those records after the stored facts change. An update can
change which older record is needed, even when the question stays the same.

Consider a long interaction history. An early record places a weekly meeting
in Room A. Separate records place Room A on floor 2 and Room B on floor 5.
Much later, an update moves the meeting to Room B. For the question
``On which floor is the weekly meeting?'', the relevant evidence is now the
move and the earlier record about Room B. Before the update, the same
question required the record about Room A. The question and both floor
records are unchanged. Only the current meeting room has changed, yet that
change determines which earlier record the assistant must read.

\begin{figure}[t]
\centering
\includegraphics[width=\linewidth]{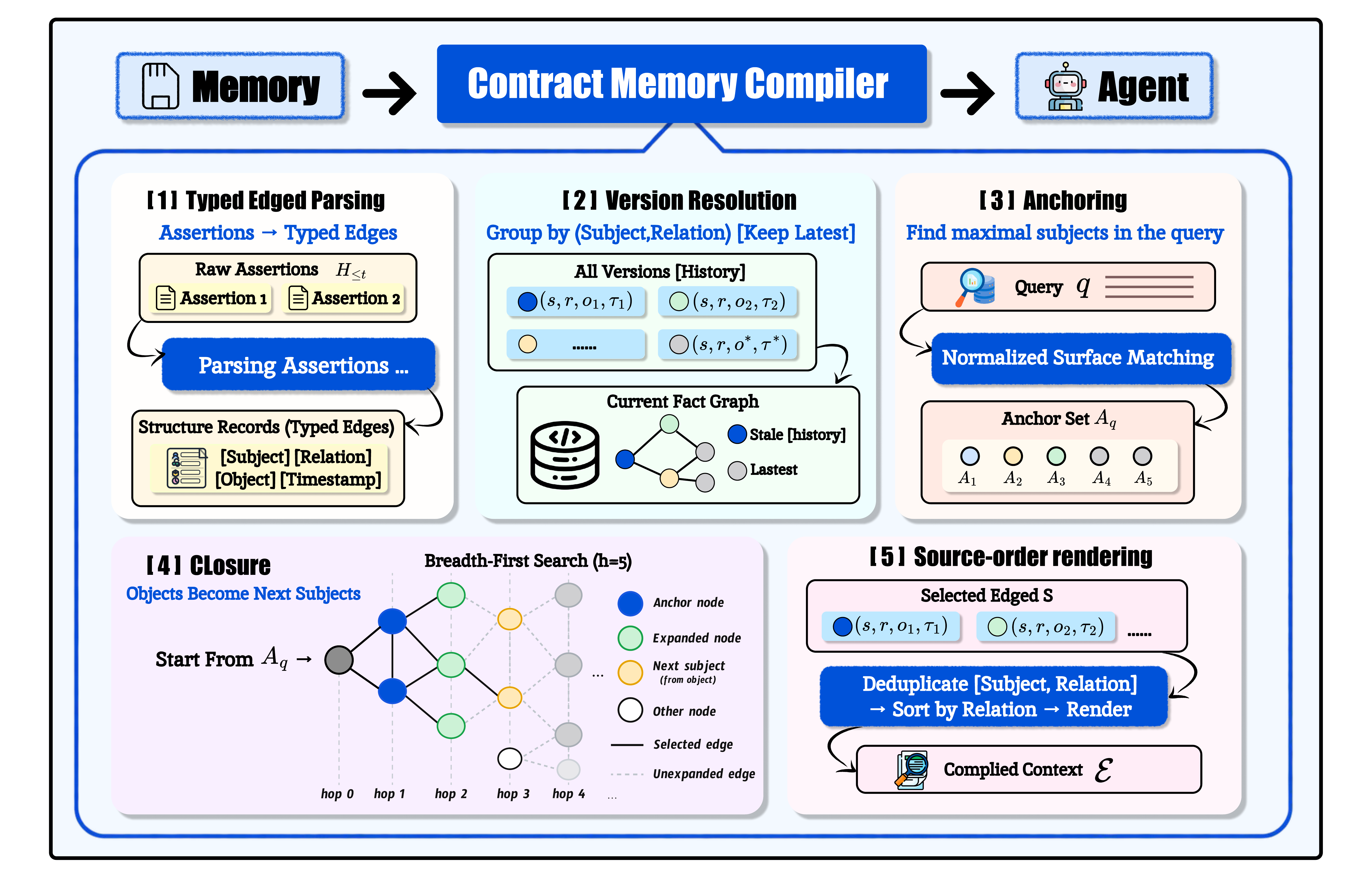}
\caption{The Contract Memory Compiler between the assertion memory and the
reader. Typed-edge parsing and version resolution build one current object
per subject--relation key. Anchoring, closure, and source-order rendering
run once per query: explicit subjects start a five-hop traversal in which
each object becomes the next subject, and the selected source records are
rendered for one reader call.}
\label{fig:cmc-pipeline}
\end{figure}

\newpage
The memory must therefore resolve the current room and use it to select the
next record. If selection precedes the update, the context can contain
Room A's floor while omitting Room B's. Refreshing the meeting-room fact
inside that context cannot recover the omitted floor record. Similarity search may
find the move but miss the floor record because Room B is absent from the
question. Iterative retrieval can issue another query after identifying
Room B \citep{trivedi2023ircot}. Our setting instead selects
evidence for one answer call. We call this task
\emph{update-dependent evidence selection}. It requires the current relation
to guide record selection before the reader receives its context.

The Contract Memory Compiler (\method{}) implements this order. It prepares the memory before
questions arrive (Figure~\ref{fig:cmc-pipeline}). It uses a language model to
extract simple relationships from the records. It stores each relationship
with the position of its source record. It then processes the relationships
in source order. For each subject and relation, it keeps the latest value.
In the example, \method{} identifies Room B as the current meeting room.
The older note about Room B's floor remains available. When the question
arrives, \method{} starts from the meeting named in the question. It follows
the current location to Room B. It then reads Room B's floor. The answer
model receives the original move and floor notes. It answers once.

We evaluate this read order on FactConsolidation, a MemoryAgentBench dataset for
selective forgetting. It has long, ordered histories with contradictory
updates. Its questions ask about the final memory state
\citep{hu2026memoryagentbench}. On questions that require several hops,
\method{} achieves 78.25\% substring exact match overall and 61.0\% on
262K token histories. To the best of our knowledge, these results establish
the state of the art for FactConsolidation multi-hop questions.

To test update-dependent selection on controlled multi-hop chains, we construct
MQuAKE-MemStream from the counterfactual cases of
MQuAKE-Remastered \citep{zhong2025mquakeremastered}. This derived dataset
turns released facts before and after an edit into ordered memory streams.
It provides a second controlled test of evidence selection after updates.

Our contributions are:
\begin{itemize}
  \item We define a bounded read contract. It specifies how updates affect current values and evidence coverage.
  \item We implement the contract in \method{}. It extracts relations before questions arrive, resolves updates in order, and selects original records for one answer call.
  \item We construct MQuAKE-MemStream, a derived dataset for multi-hop memory reading based on MQuAKE-Remastered counterfactual cases. It provides ordered histories and controls for version resolution and evidence selection.
\end{itemize}

\section{Reading a Serially Overwritten Memory}
\label{sec:problem}

\paragraph{Setting.}
Let $H_{\le t}$ be an ordered assertion stream.  A parsed assertion is an edge
$e=(s,r,o,\tau)$, where $s$ is its subject, $r$ its relation, $o$ its object,
and $\tau$ its source position.  Relations follow last-write-wins semantics:
a later assertion overwrites an earlier one with the same key $(s,r)$.
The current state contains one edge per key,
\[
 G_t = \left\{\operatorname*{arg\,max}_{e\in\mathcal{E}_t:\,
 e.\mathrm{key}=k} e.\tau : k\in\mathrm{keys}(\mathcal{E}_t)\right\},
\]
where $\mathcal{E}_t$ is the set of parsed edges.  Recency is defined by source
order in this setting.

\paragraph{The read contract.}
For a query $q$, let $A_q$ be its explicitly named graph subjects and let
$d_{G_t}(A_q,s)$ be the shortest directed distance from any anchor to $s$.
For hop limit $h$, the selected evidence is
\[
 C_h(q,G_t)=\{(s,r,o,\tau)\in G_t : d_{G_t}(A_q,s)<h\}.
\]
This gives two checkable properties of the selected interpreted edges.
\emph{Currency}: every edge in $C_h$ is the last-written parsed edge for its
key. \emph{Bounded graph coverage}: every current outgoing edge in $G_t$ from
a subject reachable at depth less than $h$ is included. We call this set a
bounded closure. It follows graph reachability and may include branches
beyond the relation requested by the query. Its usefulness depends on the
supporting facts being parsed and reachable within $h$ hops of an explicit
anchor. These graph properties do not assign versions to unresolved source
text retained alongside the selected edges.

\paragraph{Updates can change which keys a read needs.}
In the running example, the meeting moves from Room A to Room B while
both rooms keep their floors. Before the update, a read contains
meeting $\to$ Room A and Room A $\to$ floor 2. Replacing only the values of
these previously selected keys yields meeting $\to$ Room B and Room A
$\to$ floor 2: both edges are current, but Room B's floor is absent.
Recomputing reachability in $G_t$ selects meeting $\to$ Room B and Room B
$\to$ floor 5. The second lookup must change from Room A's floor to
Room B's floor. This motivates resolution before traversal. Our experiments
separately test value choice within fixed keys and key selection before value
refresh.

\paragraph{Reader interface.}
The compiler renders source records for $C_h$ together with unresolved records
as evidence $E$ for a language-model reader $R(q,E)$ under a fixed input
window. All CMC contexts in the shared direct-extraction comparison fit its
window. Selecting among competing closures when the window binds remains a
separate extension of this read interface.

\section{Contract Memory Compiler}
\label{sec:method}

\method{} takes an ordered assertion history and prepares evidence for its
reader (Figure~\ref{fig:cmc-pipeline}). Its preparation stage extracts
relations, their source positions, and unresolved original records, then
materializes a current graph shared across queries about that history.
For each question, it finds explicit subject anchors, traverses current
outgoing edges up to five hops, and renders the supporting source records.
Version resolution, anchoring, traversal, and rendering are deterministic.

\paragraph{Integration boundary and lifecycle.}
This preparation runs before query-time reading. The selected facts
are returned as original text, together with unresolved records, for one
frozen-reader generation. An empty graph selection uses the original
history under the same evidence budget. The internal loop expands a graph
frontier: each selected object identifies a subject to inspect next.
Algorithm~\ref{alg:cmc} gives the shared-LLM procedure evaluated below.

\paragraph{Stage 1: parse and retain source provenance.}
The frontend receives source records and a catalogue of relation names from
previous extraction batches. It assigns each record one status:
\texttt{facts}, containing directed subject--relation--object triples;
\texttt{unresolved}, retaining the record for direct reading; or
\texttt{no\_fact}, contributing neither an edge nor a residual. Structural
validation checks record identities, the JSON schema, and literal subject
and object spans. Source-only LLM alignment proposes a common name for
equivalent relation slots. Fixed case, punctuation, and possessive
normalization produces entity keys. Accepted edges retain the source position
and original text; unresolved records form the residual set $U$. Offline
extraction, alignment, and repair use source text alone
(Appendix~\ref{sec:llm-frontend}). The grammar adapter uses manually specified
sentence patterns for this stage.

\paragraph{Stage 2: materialize the current graph.}
Accepted assertions are replayed in source order into a dictionary keyed by
normalized subject and aligned relation. Each occurrence replaces that key's
stored edge, leaving other keys unchanged. The resulting $G_t$ is shared
across queries about the history; original assertions remain in the source.
For example, ``[4] The weekly meeting is in Room B'' produces
$(\texttt{weekly\_meeting},\texttt{held\_in},\texttt{room\_b},4)$, storing
$(\texttt{room\_b},4)$ at key $(\texttt{weekly\_meeting},\texttt{held\_in})$.
Traversal follows this value to Room B, and rendering returns record 4 and
the original record stating Room B's floor.

\paragraph{Stage 3: anchor explicit query entities.}
The compiler normalizes the question and finds graph subjects that occur as
whole normalized text spans.  If one matched subject is contained in a longer
matched subject, it keeps the longer match.  All remaining matches form $A_q$.
This is a deterministic surface-form rule; entity disambiguation and alias
inference are outside this implementation.

\paragraph{Stage 4: traverse the current graph.}
The question determines the initial subject anchors. From these anchors,
breadth-first traversal includes every current outgoing edge at subject
depths zero through four. Each selected object supplies a subject for the
next expansion, so a revised value redirects subsequent evidence selection.
The result is a bounded, potentially branching subgraph. Shortest-depth
bookkeeping handles cycles, and edges are deduplicated by $(s,r)$.
Expansion stops when the frontier is empty or the five-hop cap is reached.
Relation choice is left to the reader, which receives the selected original
assertions.

\paragraph{Stage 5: render source records.}
For a nonempty selected subgraph, CMC retrieves the original numbered
records supporting its edges and adds all unresolved source records.
An empty selection uses the original history. The renderer deduplicates
records by source position, restores source order, and applies a
60{,}000-character budget including the evidence prefix. On overflow, it
retains the newest contiguous suffix of whole records. The reader receives
the evidence and question with the instruction that larger serial numbers
are newer. All selected evidence fits in the primary shared-extraction
comparison. Currency and bounded coverage apply to the selected parsed
edges; unresolved text retains its original wording.

\paragraph{Additional memory interfaces.}
The broader MemoryAgentBench evaluation \citep{hu2026memoryagentbench}
includes tasks adapted from LongMemEval \citep{wu2024longmemeval}, ReDial
\citep{li2018redial}, $\infty$Bench \citep{zhang2024infinitebench}, and
DetectiveQA \citep{xu2024detectiveqa}, alongside document QA and
classification tasks. We use original-window retrieval for accurate
retrieval and recommendation, card/raw lookup for classification and
DetectiveQA, and coverage notes with original-text fallbacks for
summarization. The window interface
uses 2{,}048-character windows with 256-character overlap, BM25 selection,
and source-order rendering. Dataset origins and interface details are in
Appendix~\ref{app:mab-full-protocol}.

\paragraph{Implementation cost.}
Latest-state construction is one ordered pass over the frozen extraction.
A graph read scans keys for anchors, traverses the bounded subgraph, and
sorts selected records by source position. At 262K, the measured CPU read
takes 13.8\,ms median (144.3\,ms p95); Appendix~\ref{app:runtime-audit}
reports the timing protocol. Reader tokens include the chat template;
extraction cost is separate.

\section{Experiments}
\label{sec:experiments}

We first test whether resolving updates before selecting evidence improves
answers on FactConsolidation. Its 800 questions comprise equal single- and
multi-hop sets across 6K, 32K, 64K, and 262K histories; we use the released
substring exact-match score. The primary comparison fixes the extracted
facts, answer model, and prompt, changing only the records placed in the
answer context. We then vary the reader and extraction method, test
MQuAKE-MemStream, and report broader MemoryAgentBench
results. Implementation and evaluation protocols are in
Appendices~\ref{app:implementation}, \ref{app:llm-protocols},
\ref{app:mquake-stream-protocol}, and~\ref{app:mab-full-protocol}.

\subsection{FactConsolidation: selecting evidence after updates}
\label{sec:shared-llm-results}

CMC's preparation stage processes all 18{,}336 unique source statements
without seeing questions or answer labels. We freeze the extracted relations
and unresolved records for all parsed views in Table~\ref{tab:shared-llm}.
Every view uses the same source histories, HY4-preview reader, prompt, and
60{,}000-character evidence cap; only the evidence presented to the reader
changes.

\noindent
\begin{minipage}[t]{0.52\textwidth}
\vspace{0pt}
\begin{table}[H]
\small
\caption{FactConsolidation with shared extraction and one HY4-preview reader.
SH and MH are accuracy (\%); Tokens is mean reader-input length; Trunc. counts
clipped contexts.}
\label{tab:shared-llm}
\mabsetup
\setlength{\tabcolsep}{3.4pt}
\begin{tabular}{lrrrr}
\toprule
\rowcolor{mabhead}
View & SH & MH & Tokens & Trunc. \\
\midrule
\mabband{4}{mablong}{Untruncated views} \\
\rowcolor{tencentrow}
\thiswork{CMC}
  & 97.25 & \mabb{78.25} & 185 & 0 \\
Prior read + refresh
  & 97.25 & 21.50 & 195 & 0 \\
Stale closure
  & 32.00 & 11.75 & 182 & 0 \\
Fact BM25
  & \mabb{97.50} & 20.75 & 1{,}528 & 0 \\
\mabband{4}{mabrag}{Budgeted (600/800 truncated)} \\
Latest state
  & 53.00 & 17.00 & 12{,}351 & 600 \\
Raw history
  & 27.50 & 5.00 & 12{,}870 & 600 \\
\bottomrule
\end{tabular}
\end{table}
\end{minipage}

\hfill
\begin{minipage}[t]{0.45\textwidth}
\vspace{0pt}
\raggedright
Prior read + refresh chooses records using previous relation values and then
replaces the selected values with current ones. Stale closure keeps CMC's
selected relations but uses previous values. Fact BM25 ranks current facts
by question overlap. CMC and these three controls fit the input limit.
Latest state and raw history omit records on 600 questions; their scores
include truncation. Figure~\ref{fig:cmc-shared-evidence} separates history
lengths. Full protocols and extraction costs are in
Appendix~\ref{app:llm-protocols}.
\end{minipage}

\begin{figure}[H]
\centering
\includegraphics[width=\linewidth]{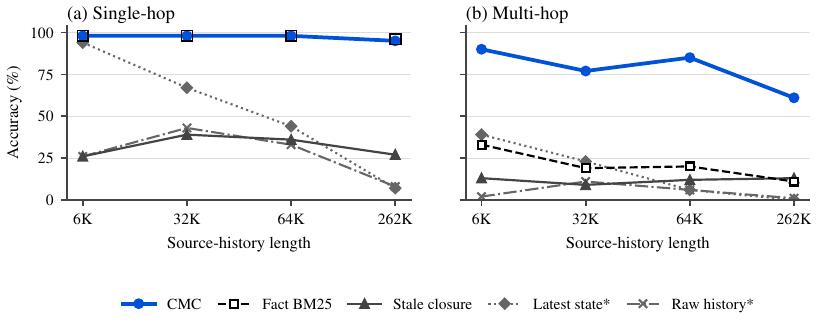}
\caption{FactConsolidation accuracy by history length. Each point covers
100 questions with the same HY4-preview reader. Latest-state and raw-history
scores include truncation on 600 of 800 questions.}
\label{fig:cmc-shared-evidence}
\end{figure}

\paragraph{Selecting after updates.}
Prior read + refresh chooses records from previous relation values and then
refreshes those records. All presented parsed facts are current, but newly
needed records can be absent. CMC ties this control at 97.25\% single-hop
and scores 78.25\% versus 21.50\% multi-hop. With extraction and reader
fixed, the 56.75-point gap isolates record selection
(Appendix~\ref{app:order-control}).

\paragraph{Retrieving current facts.}
CMC and fact BM25 are nearly tied on single-hop questions (97.25\% and
97.50\%): the BM25 pool contains the scored answer string on all 389
single-hop questions CMC answers correctly. The gap appears on multi-hop
questions (78.25\% versus 20.75\%), where an intermediate entity can be
absent from the question. In one 262K counterfactual history, the question
asks for the origin of the sport played by Christian Abbiati. CMC follows
Abbiati $\to$ cornerback $\to$ field hockey $\to$ Philippines; BM25
retrieves the Abbiati fact but misses the record giving field hockey's
origin. Across the 400 multi-hop questions, CMC is correct on 233 items
where BM25 is wrong, versus three in the other direction. On 218 of those
233 items, BM25 lacks at least one CMC-selected fact; on 109, it lacks the
scored answer string. Neither read is truncated, and CMC uses fewer mean
reader-input tokens (185 versus 1{,}528). These coverage diagnostics show
where question-based ranking often loses the newly relevant next hop.

\begin{figure}[H]
\centering
\includegraphics[width=\linewidth]{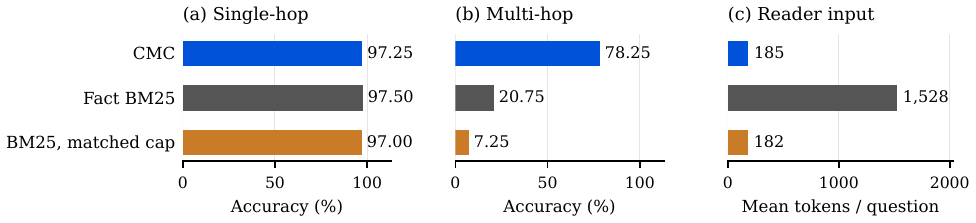}
\caption{FactConsolidation at matched evidence length. BM25 packs whole
current facts within each question's CMC budget. Panel (c) shows mean
HY4-preview reader-input tokens over all 800 questions.}
\label{fig:cmc-matched-budget}
\end{figure}

\paragraph{Evidence selection at matched length.}
We also give BM25 the same per-question evidence budget as CMC, packing
whole current facts from a fixed top-100 ranking. Mean reader inputs are
then 182 versus 185 tokens. Single-hop scores remain close (97.00\% versus
97.25\%), but BM25 reaches only 7.25\% multi-hop versus CMC's 78.25\%
(Figure~\ref{fig:cmc-matched-budget}). On the 286 multi-hop questions
answered only by CMC, the matched BM25 pool never contains every
CMC-selected fact and contains the scored answer string on just 19.
The difference persists when the reader receives nearly equal amounts of
text. Appendix~\ref{app:matched-budget} gives the packing rule and results
by history length.

Appendix Table~\ref{tab:same-fact} compares graph selection with lexical and
dense ranking over the same extracted current facts. BGE-M3 dense retrieval reaches
33.25\% multi-hop accuracy with 1{,}535 mean reader-input tokens, above fact
BM25's 20.75\% but below CMC's 78.25\% with 185 tokens. Changing the
similarity function improves retrieval, but still leaves many next-hop
records unnamed by the question.

Version choice also matters. Stale closure keeps CMC's selected relations
but substitutes previous values where available; it reaches 11.75\%
multi-hop with 182 mean input tokens. This isolates the cost of following
old values from the earlier control's cost of choosing records before
resolution. Raw history and the latest-state view reach 5.00\% and 17.00\%
under the shared cap, but both omit records on 600 questions. Their scores
therefore reflect truncation as well as evidence selection.

Traversal depth also affects coverage: multi-hop accuracy is 5.50, 54.00,
71.75, and 78.25\% at caps of one, two, three, and five hops, respectively
(Appendix~\ref{app:hop-depth}).

\paragraph{Query-time conflict resolution.}
We adapt a released post-retrieval conflict-resolution pipeline
\citep{reddy2026reliableassembly} to the same 800 histories and questions.
With HY4-preview, it reaches 98.00\% single-hop and 60.75\% multi-hop,
versus CMC's 97.25\% and 78.25\%. CMC gains 79 multi-hop answers and loses
nine. Each system retains its own extraction and call sequence, so this is
an end-to-end comparison rather than a fixed-extraction control.
Appendix~\ref{app:car-comparison} gives the adaptation and call costs.

\subsection{Reader and extraction changes}
\label{sec:reader-ablation}

We first replace the HY4-preview answer model while keeping the LLM
extraction and evidence packets fixed. With Qwen3.5-4B, CMC reaches 76.50\%
multi-hop versus 20.50\% for fact BM25; with Llama-3.1-8B, it reaches
55.00\% versus 11.50\%. Single-hop scores remain close within each reader
(Table~\ref{tab:factcon-readers}). Thus the multi-hop evidence advantage
persists across these readers, although their absolute scores differ.

We separately replace the LLM extraction with a grammar adapter. Under
Qwen3.5-4B and Qwen3.5-9B, CMC reaches 79.00\% and 80.25\% multi-hop,
versus 22.50\% for each reader given the untruncated current state. With
Qwen3.5-4B, a size-matched random negative control that excludes
CMC's selected keys scores 1.25\% multi-hop. These
rows test a different extraction and control from the reader-swap rows;
Appendix~\ref{app:complete-results} gives the full grammar grid.
With the same HY4-preview reader and CMC read, direct LLM extraction reaches
78.25\% multi-hop accuracy, close to the grammar adapter's 79.75\%, while
deriving relation names from source records rather than manually specified
patterns (Appendix~\ref{sec:llm-frontend}).

\begin{table}[ht]
\centering
\small
\caption{FactConsolidation reader/frontend settings, 800 questions each.
The grammar rows share one extraction and use untruncated current state as
control. The LLM rows share the extracted facts and original reader messages;
the control is fact BM25. Each reader uses its own chat interface.}
\label{tab:factcon-readers}
\mabsetup
\setlength{\tabcolsep}{3.6pt}
\begin{tabular}{llrrrr}
\toprule
\rowcolor{mabhead}
Reader & Frontend
  & CMC SH & CMC MH
  & Ctrl.\ SH & Ctrl.\ MH \\
\midrule
\mabband{5}{mablong}{Grammar adapter; control $=$ full current state} \\
Qwen3.5-4B & Grammar
  & 97.25 & \mabb{79.00}
  & 95.25 & 22.50 \\
Qwen3.5-9B & Grammar
  & \mabb{98.00} & \mabb{80.25}
  & 94.25 & 22.50 \\
\mabband{5}{mabrag}{Shared LLM extraction; control $=$ fact BM25} \\
Qwen3.5-4B & Shared LLM
  & 96.50 & \mabb{76.50}
  & 96.50 & 20.50 \\
Llama-3.1-8B & Shared LLM
  & 90.75 & \mabb{55.00}
  & 92.00 & 11.50 \\
\rowcolor{tencentrow}
HY4-preview & Shared LLM
  & 97.25 & \mabb{78.25}
  & \mabb{97.50} & 20.75 \\
\bottomrule
\end{tabular}
\end{table}

\subsection{MQuAKE-MemStream: a constructed memory dataset}
\label{sec:mquake-stream}

We construct MQuAKE-MemStream from the two-hop counterfactual cases of
MQuAKE-Remastered \citep{zhong2025mquakeremastered,zhong2023mquake}.
It converts released pre- and post-edit facts into ordered memory streams
while retaining the source questions and answer labels. Every method
receives the answer-bearing support facts, so this is a controlled test
of version resolution and evidence selection rather than the original
model-editing task. The transformation, pooling, and consistency audit
are in Appendix~\ref{app:mquake-stream-protocol}.

\begin{table}[ht]
\centering
\small
\caption{MQuAKE-MemStream, constructed from the two-hop MQuAKE-Remastered
cases. Frozen Qwen3.5-4B reader; 800 single-hop and 798 multi-hop questions.
Raw history uses its newest suffix; the last two rows are
percentage-point differences from CMC.}
\label{tab:mquake-stream}
\mabsetup
\setlength{\tabcolsep}{8pt}
\begin{tabular}{lrr}
\toprule
\rowcolor{mabhead}
Context given to the reader & Single hop & Multi hop \\
\midrule
\rowcolor{tencentrow}
\thiswork{CMC}
  & \mabb{99.63} & \mabb{96.62} \\
Latest state
  & 98.75 & 53.38 \\
Stale closure
  & 30.88 & 13.41 \\
Raw history (windowed)
  & 89.63 & 60.53 \\
\midrule
CMC $-$ latest state
  & $+0.88$ & $+43.23$ \\
CMC $-$ stale closure
  & $+68.75$ & $+83.21$ \\
\bottomrule
\end{tabular}
\end{table}

With the pinned Qwen3.5-4B reader (Table~\ref{tab:mquake-stream}), CMC reaches
96.62\% multi-hop accuracy, versus 53.38\% when the reader sees all current
facts and 13.41\% when the selected relations use earlier values. Selecting
a connected read from the current state gains 43.23 points over presenting
that state in full. Resolving values within the selected relations gains
83.21 points over stale values.

\subsection{Same-reader MemoryAgentBench}
\label{sec:mab-full-results}

Table~\ref{tab:mab-hy4-main} reports the full MemoryAgentBench grid
\citep{hu2026memoryagentbench}. Only SF uses the current-graph read;
direct comparisons use the four local HY4-preview rows.

\definecolor{tencentblue}{HTML}{0052D9}
\definecolor{tencentwash}{HTML}{E7F1FC}
\definecolor{tencentrow}{HTML}{C5D8F7}
\definecolor{mabhead}{HTML}{F3F3F3}
\definecolor{mablong}{HTML}{E6E2D8}
\definecolor{mabrag}{HTML}{F2E6D4}
\definecolor{mabagent}{HTML}{EADFE6}
\begingroup
\providecommand{\mabb}[1]{\textbf{#1}}
\providecommand{\mabu}[1]{\underline{#1}}
\newcommand{\mabgrp}[2]{%
  \cellcolor{#1}\rule[-2.6pt]{0pt}{11pt}%
  & \multicolumn{10}{c}{\cellcolor{#1}{\itshape #2}}}
\newcommand{\mabgrptx}[1]{%
  \cellcolor{tencentblue}\rule[-2.6pt]{0pt}{11pt}%
  & \multicolumn{10}{c}{\cellcolor{tencentblue}{\color{white}\itshape\bfseries #1}}}

\begin{table*}[t]
\centering
\caption{MemoryAgentBench. Upper blocks copy published Table~3 numbers as
reference; their readers differ and they are not bolded against this work.
The last block holds one HY4-preview reader. AR and Rec are original-window
retrieval. SF is current-graph traversal on overwritten
assertions; the two cells are the official 262K split (200 questions).
MCC, Summary, and DetectiveQA are the original same-reader campaign, not
Algorithm~\ref{alg:cmc}: card/raw lookup on MCC and DetectiveQA, coverage
notes on Summary, and the registered JSON-field DetQA decoder. Bold is the
highest value \emph{inside the last block} of each column.}
\label{tab:mab-hy4-main}
\setlength{\tabcolsep}{4.2pt}
\renewcommand{\arraystretch}{1.12}
\setlength{\aboverulesep}{0pt}
\setlength{\belowrulesep}{0pt}
\resizebox{\textwidth}{!}{%
\begin{tabular}{lcccccccccc}
\toprule
& \multicolumn{4}{c}{\textbf{AR: original windows}}
& \multicolumn{1}{c}{\textit{MCC}}
& \multicolumn{1}{c}{\textbf{Rec: windows}}
& \multicolumn{2}{c}{\textit{LRU}}
& \multicolumn{2}{c}{\textbf{SF: current graph}} \\
\rowcolor{mabhead}
\textbf{System}
& SH-QA & MH-QA & LME(S*) & EventQA
& MCC & Rec.
& Sum & DetQA
& FC-SH & FC-MH \\
\midrule
\mabgrp{mablong}{Published long-context (reference)} \\
GPT-5-mini (400K)
  & 85.0 & 71.0 & 63.3 & 78.2
  & 84.0 & 13.2
  & 56.3 & 76.1
  & 78.0 & 28.0 \\
\mabgrp{mabrag}{Published retrieval (reference)} \\
BM25
  & 66.0 & 56.0 & 45.3 & 74.6
  & 75.4 & 13.6
  & 19.0 & 52.1
  & 48.0 & 3.0 \\
Text-Embed-3-Small
  & 60.0 & 44.0 & 48.3 & 63.0
  & 70.0 & 15.3
  & 17.7 & 54.9
  & 28.0 & 3.0 \\
Text-Embed-3-Large
  & 54.0 & 44.0 & 50.3 & 70.0
  & 72.4 & 16.2
  & 18.2 & 56.3
  & 28.0 & 4.0 \\
Qwen3-Embedding-4B
  & 57.0 & 47.0 & 43.3 & 71.4
  & 78.0 & 12.2
  & 14.8 & 59.2
  & 29.0 & 3.0 \\
RAPTOR
  & 29.0 & 38.0 & 34.3 & 45.8
  & 59.4 & 12.3
  & 13.4 & 42.3
  & 14.0 & 1.0 \\
GraphRAG
  & 47.0 & 47.0 & 35.0 & 34.4
  & 39.8 & 9.8
  & 0.4 & 39.4
  & 14.0 & 2.0 \\
MemoRAG
  & 29.0 & 33.0 & 20.0 & 56.0
  & 77.0 & 13.1
  & 9.2 & 50.7
  & 21.0 & 7.0 \\
HippoRAG-v2
  & 76.0 & 66.0 & 50.7 & 67.6
  & 61.4 & 10.2
  & 14.6 & 57.7
  & 54.0 & 5.0 \\
Mem0
  & 25.0 & 32.0 & 36.0 & 37.5
  & 32.4 & 10.0
  & 4.8 & 36.6
  & 18.0 & 2.0 \\
Cognee
  & 31.0 & 26.0 & 29.3 & 26.8
  & 35.4 & 10.1
  & 2.3 & 29.6
  & 28.0 & 3.0 \\
Zep
  & 44.0 & 25.0 & 38.3 & 42.5
  & 62.8 & 12.1
  & 4.2 & 28.2
  & 7.0 & 3.0 \\
Self-RAG
  & 35.0 & 42.0 & 25.7 & 31.8
  & 11.6 & 12.8
  & 0.9 & 35.2
  & 19.0 & 3.0 \\
MemGPT
  & 41.0 & 38.0 & 32.0 & 26.2
  & 67.6 & 14.0
  & 2.5 & 42.3
  & 28.0 & 3.0 \\
MIRIX
  & 62.0 & 61.0 & 37.3 & 29.8
  & 38.4 & 9.8
  & 9.9 & 40.8
  & 14.0 & 2.0 \\
MIRIX (4.1-mini)
  & 73.0 & 75.0 & 51.0 & 53.0
  & 61.0 & 10.3
  & 18.9 & 62.0
  & 20.0 & 3.0 \\
\mabgrptx{Same HY4-preview reader} \\
\rowcolor{tencentwash}
HY4-preview (suffix)
  & 18.0 & 19.0 & 6.0 & 75.4
  & \mabb{86.4} & 13.1
  & 14.3 & 76.1
  & 8.0 & 1.0 \\
HY4 BM25
  & 78.0 & 60.0 & 39.3 & 85.6
  & 85.8 & 14.1
  & 14.8 & 77.5
  & 30.0 & 2.0 \\
HY4 dense+BM25
  & 69.0 & 44.0 & 37.3 & 87.2
  & 86.2 & 11.9
  & \mabb{15.2} & \mabb{78.9}
  & 28.0 & 1.0 \\
\rowcolor{tencentrow}
\textbf{\textcolor{tencentblue}{CMC + task interfaces}}
  & \mabb{85.0} & \mabb{77.0} & \mabb{44.3} & \mabb{90.8}
  & 86.2 & \mabb{17.8}
  & 10.2 & 67.6
  & \mabb{95.0} & \mabb{61.0} \\
\bottomrule
\end{tabular}}
\end{table*}
\endgroup

At the 262K FactConsolidation split, CMC reaches 95.0\% single-hop and
61.0\% multi-hop, versus 8.0\% and 1.0\% for the suffix and 30.0\% and
2.0\% for same-reader BM25. These cells are a subset of the 800-question
comparison above.

CMC leads the same-reader rows on five retrieval and recommendation columns and
is similar on classification (85.8--86.4), but trails dense+BM25 on
summarization (10.2 versus 15.2) and DetectiveQA (67.6 versus 78.9).
Task protocols and diagnostics are in Appendix~\ref{app:mab-full-protocol}.

\section{Related Work}

\paragraph{Writing and maintaining agent memory.}
Memory Networks and RAG distinguish stored material from what is read for a
task \citep{weston2015memory,miller2016kvmem,lewis2020rag}. Long-term dialogue
adds multi-session recall and invalidation
\citep{xu2022goldfish,bae2022keepupdated}. Generative Agents retrieve
experiences, and MemGPT moves context across memory tiers
\citep{park2023generativeagents,packer2023memgpt}. SimpleMem and Mem0 compress
dialogue \citep{liu2026simplemem,chhikara2025mem0}; A-Mem revises linked notes
\citep{xu2025amem}. RMM builds summaries at multiple scales and refines
retrieval using cited evidence \citep{tan2025rmm}. AgeMem learns storage and
update actions \citep{yu2026agemem}; THEANINE retains older events in
timelines \citep{ong2025theanine}. Our question is which original records enter
one answer context after a current relation changes.

\paragraph{Current state and graph retrieval.}
APEX-MEM organizes temporally grounded events in a property graph and uses
an agent to resolve changing information at query time
\citep{banerjee2026apexmem}. MemTxn checks whether writes have source support
and selects visible versions of conflicting facts \citep{cui2026memtxn}.
Zep and SodaMem represent changing relationships in temporal graphs
\citep{rasmussen2025zep,wan2026sodamem}.
GAM separates event progression from longer-term topic consolidation
\citep{wu2026gam}. MAGMA traverses semantic, temporal, causal, and entity
graphs \citep{jiang2026magma}. HippoRAG uses a knowledge graph and
personalized PageRank for multi-hop retrieval \citep{gutierrez2024hipporag};
Mnemis combines similarity search with hierarchical graph selection
\citep{tang2026mnemis}. GraphRAG summarizes graph communities for corpus-level
questions; RAPTOR retrieves through a summary tree
\citep{edge2024graphrag,sarthi2024raptor}. ReFind searches raw dialogue
turns \citep{li2026refind}.
A stage-wise comparison of graph and non-graph memories likewise
finds that implementation choices affect measured performance
\citep{hu2026memorygraphs}. We isolate a narrower property: under
single-valued, ordered assertions, the resolved object determines which
source record a bounded traversal selects next.

\paragraph{Multi-hop evidence and answer assembly.}
IRCoT interleaves reasoning steps with new retrieval requests when later
evidence depends on earlier findings \citep{trivedi2023ircot}. MeLLo handles
multi-hop questions after knowledge edits, while MQuAKE-Remastered studies
how edits affect downstream answers
\citep{zhong2023mquake,zhong2025mquakeremastered}. SCM routes queries among
lexical, graph, and iterative memory reads \citep{tobkin2026scm}. A
post-retrieval assembly method separates evidence extraction from the final
answer policy \citep{reddy2026reliableassembly}. CMC studies a different
boundary: it resolves updates and selects source records before a single
answer call. Our fixed-extraction controls compare this order of operations;
they do not rank complete adaptive pipelines against one another. The
G-Walk and MeLLo ports in Appendix~\ref{app:same-reader-ports}
receive a resolved fact bank, so
that comparison holds version choice fixed. MQuAKE-MemStream reads supplied
support facts rather than measuring parametric editing.

\section{Discussion}

The multi-hop gain comes from selecting connected current facts: refreshing
previously selected values loses 56.75 points
(Section~\ref{sec:shared-llm-results}). The LLM frontend avoids hand-written
patterns, but exact anchoring, missing links, and the hop limit restrict
coverage. FactConsolidation came from a broader development grid, so
independent confirmation remains needed. An automated screen flags 33 of
800 reference answers as possibly superseded; all remain in the official
scores. The multi-hop gap persists under strict normalized exact match and
is at least 50.50 points under adversarial rescoring of the 26 flagged
multi-hop items (Appendix~\ref{app:label-sensitivity}).

\section{Conclusion}

\method{} resolves current relations before it selects evidence.
On ordered, single-valued assertions, that order lets an overwrite redirect
the next hop: refreshing the previously selected graph reaches 21.50\%
multi-hop accuracy, and traversing the current graph reaches 78.25\%.
The same gap appears against retrieval over the current facts and under two
additional readers. The demonstrated gain is for ordered, single-valued
assertion streams.

\section*{Reproducibility Statement}

We record inputs, model identifiers, prompts, settings, outputs, and native
scores, and verify reused responses before joint scoring.
Appendix~\ref{app:llm-protocols} documents the evaluation protocol.
The supplementary package includes the compiler, evaluation scripts,
MQuAKE-MemStream data, and per-example predictions.

\section*{AI Use Statement}

AI coding assistants assisted implementation, experiment execution, artifact
checks, and manuscript editing. The authors retain responsibility for the
methods, results, citations, and final manuscript.

\clearpage
\bibliographystyle{plainnat}
\bibliography{references}

\clearpage
\appendix
\pdfbookmark[0]{Appendix Contents}{appendix-contents}
\section*{Appendix Contents}
\begingroup
\small
\setlength{\parskip}{2pt}
\newcommand{\appendixcontentsline}[2]{%
  \noindent\hspace*{#1}%
  \hyperref[#2]{\ref*{#2}\quad\nameref*{#2}}%
  \dotfill\pageref*{#2}\par}
\appendixcontentsline{0pt}{app:cmc-algorithm}
\appendixcontentsline{0pt}{app:complete-results}
\appendixcontentsline{1.5em}{app:grammar-grid}
\appendixcontentsline{1.5em}{app:same-reader-ports}
\appendixcontentsline{1.5em}{app:label-sensitivity}
\appendixcontentsline{0pt}{app:implementation}
\appendixcontentsline{0pt}{app:llm-protocols}
\appendixcontentsline{1.5em}{sec:llm-frontend}
\appendixcontentsline{1.5em}{app:shared-extraction-protocol}
\appendixcontentsline{1.5em}{app:shared-qwen-reader}
\appendixcontentsline{1.5em}{app:car-comparison}
\appendixcontentsline{0pt}{app:matched-budget}
\appendixcontentsline{0pt}{app:mquake-stream-protocol}
\appendixcontentsline{0pt}{app:order-control}
\appendixcontentsline{0pt}{app:hop-depth}
\appendixcontentsline{0pt}{app:mab-full-protocol}
\appendixcontentsline{0pt}{app:runtime-audit}
\endgroup
\medskip

\section{Compiler Procedure}
\label{app:cmc-algorithm}

Algorithm~\ref{alg:cmc} specifies the shared-LLM FactConsolidation read.
Source interpretation is frozen across its evidence-view comparisons.
Original-window retrieval and card interfaces in the broader benchmark
are specified in Appendix~\ref{app:mab-full-protocol}.

\begin{algorithm}[H]
\caption{\method{}: compile evidence from current assertions}
\label{alg:cmc}
\begin{algorithmic}[1]
\STATE \textbf{input} ordered source records $H_{\le t}$, query $q$, budget $B=60{,}000$
\STATE \textbf{source preparation} (shared across questions)
\STATE $\mathcal{E}, U \leftarrow \textsc{LLMParseWithSourcePositions}(H_{\le t})$
\STATE $G_t \leftarrow \textsc{LatestEdgePerKey}(\mathcal{E})$
\STATE \textbf{query-time compilation}, hop cap $h=5$
\STATE $A_q \leftarrow \textsc{MaximalExplicitSubjects}(q,G_t)$
\STATE $S\leftarrow\textsc{BoundedBFS}(G_t,A_q,h)$
\IF{$S=\emptyset$}
  \STATE $V \leftarrow H_{\le t}$
\ELSE
  \STATE $V \leftarrow \textsc{SourceRecords}(S) \cup U$
\ENDIF
\STATE $V \leftarrow \textsc{DeduplicateByPosition}(V)$
\STATE $V \leftarrow \textsc{NewestWholeRecordSuffix}(V,B)$
\STATE \textbf{return} $\textsc{RenderInSourceOrder}(V)$
\end{algorithmic}
\end{algorithm}

The budget includes the fixed evidence prefix and record serial numbers.
The suffix operation visits records from newest to oldest and stops at the
first record that does not fit. When all selected records fit, it preserves
the complete selected graph evidence and unresolved records. The reader
receives this evidence and the question in one call. Source records marked
\texttt{no\_fact} contribute neither an edge nor an unresolved record.

\section{Complete Grammar-Adapter Results}
\label{app:complete-results}

This appendix reports the complete grammar-adapter FactConsolidation arms
and implementation details. All values use the released substring exact-match
scorer and include every evaluated question in the denominator.

The separate HY4 frontend comparison appears in
Appendix~\ref{sec:llm-frontend}. Protocols for the shared-extraction controls
are specified in Appendix~\ref{app:llm-protocols}.

\subsection{Complete ablation grid}
\label{app:grammar-grid}

Table~\ref{tab:app-factcon-complete} lists seven diagnostic arms and two
same-reader method ports. Two diagnostic arms roll versions back, and they
differ in what they roll back. \emph{Stale
closure} takes the closure our compiler admitted and substitutes the
immediately preceding version when one exists, perturbing version choice while
holding selected keys and cardinality identical to ours.  Keys without a prior
version retain their active value. It is not a uniformly stale context.
\emph{Stale state}
rolls the entire query-independent current view back one version, so it
satisfies neither factor and serves as a lower reference.  The two are
distinct arms with distinct numbers and are never pooled.

\begin{table}[t]
\centering
\small
\caption{Complete grammar-adapter FactConsolidation results under one frozen
Qwen3.5-4B reader, pooled across 800 questions. Mean tokens count the full
reader input; dashes mark arms without a recorded cost total. The disjoint
random arm is a negative control: its current-state keys exclude CMC's
selected keys. G-Walk and MeLLo use a resolved one-value-per-key bank.}
\label{tab:app-factcon-complete}
\mabsetup
\setlength{\tabcolsep}{4.5pt}
\begin{tabular}{lrrr}
\toprule
& \multicolumn{2}{c}{\textbf{Accuracy (\%)}} & \\
\rowcolor{mabhead}
Context & FC-SH & FC-MH & Mean tokens \\
\midrule
\rowcolor{tencentrow}
\thiswork{CMC} & \mabb{97.25} & \mabb{79.00} & \mabb{183} \\
Latest state & 95.25 & 22.50 & 71{,}316 \\
Stale closure & 31.50 & 15.50 & --- \\
Raw history & 42.25 & 3.75 & 97{,}139 \\
Dense+reranker top-10 & 41.25 & 3.75 & 5{,}559 \\
Stale state & 31.00 & 2.50 & --- \\
Disjoint random & 9.00 & 1.25 & --- \\
G-Walk (ported) & 58.0 & 30.5 & --- \\
MeLLo (ported) & 21.5 & 8.0 & --- \\
\bottomrule
\end{tabular}
\end{table}

Restricting the four diagnostic contexts to the 600 non-overflow questions
(6K--64K) gives 97.67, 98.00, 33.33, and 38.67 single-hop accuracy for
\method{}, latest state, stale closure, and raw history, respectively; the
descriptive interaction contrast is 5.00 points.  Multi-hop accuracy is 85.67,
27.00, 15.67, and 3.33, giving a 46.33-point contrast.  The large multi-hop
non-additivity therefore remains after excluding every window-truncated
raw-history prompt.

\begin{table}[t]
\centering
\small
\caption{Grammar-adapter arms on the non-overflow 6K--64K histories under
Qwen3.5-4B, with 100 questions per cell. The 262K raw-history prompts are
truncated and are therefore omitted from this comparison.}
\label{tab:app-factcon-bylength}
\mabsetup
\setlength{\tabcolsep}{3.6pt}
\begin{tabular}{lrrrrrr}
\toprule
& \multicolumn{3}{c}{\textbf{FactCon-SH}}
& \multicolumn{3}{c}{\textbf{FactCon-MH}} \\
\rowcolor{mabhead}
Context & 6K & 32K & 64K & 6K & 32K & 64K \\
\midrule
\rowcolor{tencentrow}
\thiswork{CMC}
  & 98 & 97 & \mabb{98}
  & \mabb{91} & \mabb{81} & \mabb{85} \\
Latest state
  & \mabb{100} & \mabb{96} & \mabb{98}
  & 43 & 23 & 15 \\
Stale closure
  & 26 & 38 & 36
  & 19 & 18 & 10 \\
Raw history
  & 31 & 40 & 45
  & 2 & 0 & 8 \\
Dense+reranker top-10
  & 37 & 44 & 46
  & 3 & 2 & 8 \\
Stale state
  & 26 & 37 & 36
  & 2 & 2 & 4 \\
Disjoint random
  & 8 & 11 & 13
  & 2 & 0 & 2 \\
\bottomrule
\end{tabular}
\end{table}

\paragraph{Paired contrasts.}
Contrasts against the two controls were preregistered before the reader was
loaded and are reported as paired win, tie, and loss counts over the 800
questions with query-cluster percentile bootstrap intervals at 10,000
replicates. Against the disjoint random negative control our arm wins 667, ties
130, and loses 3, for a mean paired advantage of 83.00 points, interval
[80.25, 85.63].  Against the stale-closure arm it wins 547, ties 223, and loses
30, for a mean paired advantage of 64.63 points, interval [60.75, 68.38].  The
stale-closure arm in turn beats the negative control by 18.38 points, interval
[15.63, 21.25], which confirms that the closure structure carries information
even when its versions are wrong.

\subsection{Same-reader ports}
\label{app:same-reader-ports}

The G-Walk and MeLLo rows in Table~\ref{tab:app-factcon-complete} use the
released MQuAKE-Remastered implementations and prompts
\citep{zhong2025mquakeremastered,zhong2023mquake} with the frozen Qwen3.5-4B
reader and official FactConsolidation scorer. Their interfaces require one
object per subject--relation key, so we supply the version-resolved bank;
they do not receive the answer or its path. Published MemoryAgentBench
systems on the 262K FactConsolidation cells appear in
Table~\ref{tab:mab-hy4-main}; we do not repeat that field here.

\subsection{Scoring and label sensitivity}
\label{app:label-sensitivity}

All FactConsolidation headline scores use the released answers and
substring exact match.
Requiring each saved parsed answer to equal a reference alias after the
released normalization gives CMC, prior read + refresh, and fact BM25
77.25\%, 19.50\%, and 13.00\% on the 400 multi-hop questions, versus
official scores of 78.25\%, 21.50\%, and 20.75\%. Strict single-hop
counts are 389, 389, and 388 of 400; we do not rank these near ties.

A string-and-key screen of grammar-parsed histories flags seven single-hop
and 26 multi-hop reference strings found as earlier values of a key
reachable from the query but absent from its selected current values.
These flags need manual adjudication. On the other 374 multi-hop questions,
CMC, prior read + refresh, and fact BM25 answer 313, 85, and 83 correctly.
Assigning all 26 flagged multi-hop items the least favorable paired scores
for CMC gives full-set lower bounds of $(313-85-26)/400=50.50$ points
over prior read + refresh and $(313-83-26)/400=51.00$ points over fact
BM25. The official gaps are 56.75 and 57.50 points; this conditional
bound does not establish absolute accuracy.

\section{Implementation and Evaluation Details}
\label{app:implementation}

\paragraph{Comparability.}
Matched tables use one frozen open reader, one prompt and one decoding recipe,
and the same immutable history for every arm.  Baselines may read any of that
history; we do not subsample their evidence at random or instruct them to
ignore records outside a privileged set.  The retrieval baseline uses
non-overlapping 512-token chunks, Qwen3-Embedding-8B dense top-100 retrieval,
Qwen3-Reranker-4B top-10 reranking, and restoration of source order.  Latest
state is the strongest version-only baseline available, not an uninformed dump.
Public cross-reader rows in Table~\ref{tab:mab-hy4-main} are task
frontiers and are never used as matched ablations.

\paragraph{Frozen reader.}
Qwen3.5-4B at a pinned revision, greedy decoding, a 32-token generation cap, a
262,144-token input cap, one fixed prompt, and one output parser.  Across all
seven diagnostic arms and 5,600 generations, every output parsed and none was
empty.
Every evaluated example remains in the denominator.  All 200 raw-history
prompts in the 262K cells exceed the cap; their oldest context tokens are
left-truncated while the system instruction, query, and answer suffix are
preserved.  The 600 questions in the 6K--64K cells form the non-overflow
population.  The complete-grid values are descriptive, and
Table~\ref{tab:app-factcon-bylength} exposes the non-overflow cells separately.
The Qwen3.5-9B reader in Table~\ref{tab:factcon-readers} repeats the four
diagnostic contexts on the same 800 questions and prompt with its own frozen
reader configuration.

\paragraph{Compiler.}
The adapter recognizes the released relation grammar, selects the active edge
for each subject--relation key by largest stream position, anchors the longest
query-explicit entities, and traverses the active graph for at most five hops.
Admitted edges are deduplicated and serialized in stream order. In this
grammar-adapter grid, incomplete parsing falls back to raw history, while
an empty graph closure with complete parsing falls back to the latest state.
This differs from the direct-LLM procedure in Algorithm~\ref{alg:cmc}, where
an empty selection uses the original history. In this grid, 796 questions
have one explicit anchor and four
have two disjoint anchor closures.

\paragraph{Arm construction.}
The stale-closure arm reuses our arm's selected keys exactly and substitutes
the immediately preceding version where one exists, 3,839 substitutions over
800 questions, falling back to the active version otherwise. The disjoint
random negative control draws from the active state while excluding the
query projection keys,
matching cardinality exactly on all 800 questions and token count exactly on
797, with a mean absolute token delta of 0.005 and a maximum of 2.

\paragraph{Scoring.}
Official substring exact match from the released scorer at a pinned repository
commit against a clean worktree.  Arms, contrasts, and populations were frozen
and preregistered before the reader was loaded, and no gold value was read
before the prediction bundle was validated.  Confidence intervals on single
arms are Wilson at the 95\% level; paired contrasts use query-cluster
percentile bootstrap at 10,000 replicates.  We report intervals and paired win,
tie, and loss counts rather than significance tests.

\paragraph{Cost accounting.}
Token counts are materialized reader input including the frozen chat template.
They exclude offline parsing, indexing, embedding, graph construction, and
storage, which are paid once per write rather than once per read.

\section{Frozen LLM Evaluation Protocols}
\label{app:llm-protocols}

\subsection{Alternative Source Frontends on FactConsolidation}
\label{sec:llm-frontend}

\paragraph{Shared reader, alternative frontends.}
We compare three frontends on all 800 FactConsolidation questions using the
same HY4-preview reader, reader prompt, temperature zero, and 64-token output
cap.  The grammar adapter uses manually specified relation patterns.
The direct LLM frontend extracts all 18,336 unique source statements; its
open-vocabulary relation names are aligned using source examples.  The hybrid
frontend directly extracts 3,342 statements and induces 27 literal matching
rules from consistent source examples, using direct extraction where needed.
Neither LLM extraction nor rule induction receives questions or answer labels.
The source interpretation is adaptive to the supplied histories; it is not a
held-out test of unseen sentence forms.  Readers receive selected original
statements, with unresolved statements retained as residual evidence.

\begin{table}[t]
\centering
\small
\caption{Source-extraction frontends with the same CMC read, HY4-preview
reader, and 800 FactConsolidation questions. The grammar adapter uses
manually specified patterns; direct LLM extraction names relations from
source records; the hybrid induces literal patterns from source examples.
Column names follow MemoryAgentBench Table~3. Each hop class contains 400
questions. Scores use the official substring scorer, including failed
requests as incorrect.}
\label{tab:llm-frontend}
\mabsetup
\begin{tabular}{lrrr}
\toprule
& \multicolumn{2}{c}{\textbf{SF}} & \\
\rowcolor{mabhead}
Frontend & FC-SH & FC-MH & All \\
\midrule
Grammar adapter
  & \mabb{97.50} & \mabb{79.75} & \mabb{88.625} \\
\rowcolor{tencentrow}
\thiswork{Direct LLM}
  & 97.25 & 78.25 & 87.750 \\
Source-induced hybrid
  & 97.00 & 76.50 & 86.750 \\
\bottomrule
\end{tabular}
\end{table}

Direct LLM extraction answers 702 of 800 questions correctly, versus 709
for the manually specified grammar adapter and 694 for the source-induced
hybrid. Its multi-hop accuracy is 78.25\%, compared with 79.75\% for the
grammar adapter. This small gap shows that the observed graph-read
performance does not depend on manually specifying relation patterns for
this corpus: the direct frontend derives relation names from source records.
It does not establish generalization to unseen writing styles or relation
types. The hybrid has one failed reader request after a no-anchor read
falls back to an oversized history; the failure remains in the denominator.
We use direct LLM extraction as the primary assertion frontend and measure
the maintenance and read mechanism separately on a shared extraction.
The grammar adapter provides controlled mechanism evidence; the induced hybrid
tests reuse of source-derived patterns.
These HY4 results do not inherit the Qwen grammar experiment's token costs.

\paragraph{Source construction and freeze.}
Extraction operates on batches of source records and returns structured
statuses and triples. Validation checks the record identities, schema, and
literal subject and object spans. A source-only repair can correct malformed
extractions using the same records. Relation alignment uses source examples
and maps equivalent names to existing catalogue names; conflict review
revisits source extractions whose induced patterns disagree. The resulting
direct cache contains 18,333 one-triple records and three unresolved records.
Every source occurrence in the direct experiment is served by its cached LLM
extraction. The cache is frozen before the shared-extraction comparison, so
all five evidence views use the same interpreted facts and residuals.

The extraction prompt requests single-valued asserted slots and gives generic
direction preferences for person--employer, work--author, and
organization--location relations. It instructs the LLM to mark contextual
pronouns, negations, ambiguous scope, and simultaneous values as
\texttt{unresolved}. The LLM supplies relation names and their source-based
alignment; literal-span validation checks where the endpoints occur.

\paragraph{Extraction cost.}
The direct-extraction experiments used 827 calls, 919,509 input tokens, and
419,813 output tokens. The hybrid experiments used 178 calls, 224,251 input
tokens, and 84,716 output tokens. These totals include extraction, alignment,
and revision work, so they do not measure steady-state serving cost. The hybrid
reduces recorded extraction work at a lower observed answer accuracy.

\subsection{Shared Extraction and Natural-Dialogue Read Protocols}
\label{app:shared-extraction-protocol}

\paragraph{Shared extraction on FactConsolidation.}
The five-view protocol fixes the existing direct-LLM source extraction before
constructing CMC, latest-state, stale-closure, raw-history, and fact-BM25
contexts for all 800 questions. Structured views share the same extracted
facts and unresolved source records. Fact BM25 retrieves the top 100 active
facts using their original text, entities, and relation names. CMC traverses
up to five hops from explicit question anchors. Stale closure substitutes
the preceding observation within CMC's selected keys when one exists.
Each view restores source order and applies a 60,000-character evidence cap,
retaining a contiguous newest suffix of whole records on overflow. Thus the
budgeted latest-state view can omit active facts. This protocol measures
read construction under a common window; the Qwen grammar experiment provides
the separate untruncated current-state comparison.

All views use HY4-preview, temperature zero, no-thinking mode, a 64-token
output cap, and the same reader prompt. The 800 CMC contexts are checked
against the earlier direct-LLM experiment. Its predictions are reused only
after matching the input packet, individual inputs, reader configuration,
and prompt; the four controls require 3,200 additional requests. Scoring
begins after all five views terminate. Reader usage excludes the inherited
extraction work reported in Appendix~\ref{sec:llm-frontend}.

Table~\ref{tab:shared-llm} in the main text aggregates all 800 questions. The
original MemoryAgentBench overall table uses only the 262K SF cells (200
questions); Table~\ref{tab:shared-llm-length} separates the four history
lengths.

\begin{table}[t]
\centering
\small
\caption{Shared direct-LLM controls by source-history length, following
MemoryAgentBench Table~5 (FactCon-SH / FactCon-MH $\times$ length). Each cell
is accuracy over 100 questions with the same HY4 reader. The 262K columns are
the original overall-table SF population; Table~\ref{tab:shared-llm} pools all
four lengths.}
\label{tab:shared-llm-length}
\mabsetup
\setlength{\tabcolsep}{3.2pt}
\begin{tabular}{lrrrrrrrr}
\toprule
& \multicolumn{4}{c}{\textbf{FactCon-SH}}
& \multicolumn{4}{c}{\textbf{FactCon-MH}} \\
\rowcolor{mabhead}
View & 6K & 32K & 64K & 262K & 6K & 32K & 64K & 262K \\
\midrule
\rowcolor{tencentrow}
\thiswork{CMC}
  & 98 & \mabb{98} & \mabb{98} & 95
  & \mabb{90} & \mabb{77} & \mabb{85} & \mabb{61} \\
Latest state
  & 94 & 67 & 44 & 7
  & 39 & 23 & 6 & 0 \\
Stale closure
  & 26 & 39 & 36 & 27
  & 13 & 9 & 12 & 13 \\
Raw history
  & 26 & 43 & 33 & 8
  & 2 & 11 & 6 & 1 \\
Fact BM25
  & \mabb{98} & \mabb{98} & \mabb{98} & \mabb{96}
  & 33 & 19 & 20 & 11 \\
\bottomrule
\end{tabular}
\end{table}

\begin{table}[t]
\centering
\small
\caption{Shared LLM extraction and HY4-preview on all 800 FactConsolidation
questions. Each view selects from the same current facts and unresolved
records. Dense retrieval uses BGE-M3 top-100. Tokens are mean reader-input
tokens; Trunc. counts reads clipped by the 60,000-character rendering limit.}
\label{tab:same-fact}
\mabsetup
\setlength{\tabcolsep}{4pt}
\begin{tabular}{lrrrr}
\toprule
\rowcolor{mabhead}
View & SH & MH & Tokens & Trunc. \\
\midrule
\rowcolor{tencentrow}
\thiswork{CMC}
  & 97.25 & \mabb{78.25} & 185 & 0 \\
Fact BM25
  & \mabb{97.50} & 20.75 & 1{,}528 & 0 \\
Fact BM25, size matched
  & 97.00 & 7.25 & 182 & 0 \\
Fact dense (BGE-M3)
  & 96.50 & 33.25 & 1{,}535 & 0 \\
\bottomrule
\end{tabular}
\end{table}

\paragraph{Execution protocol.}
Transport errors may be retried; answer content and evaluation scores never
trigger retries. Terminal failures remain in the denominator. HY4-preview is
a provider alias rather than a pinned weight revision; we record the model
identifier returned for each request.

\subsection{Reader Swap with Fixed LLM Extraction}
\label{app:shared-qwen-reader}

We reuse the frozen shared-LLM comparison's exact question, evidence, and
system/user message text for CMC, fact BM25, and stale closure. Each arm
contains the same 800 questions. Only the reader and its chat template
change: Qwen3.5-4B, revision
\texttt{851bf6e806efd8d0a36b00ddf55e13ccb7b8cd0a}, runs in FP16 with SDPA,
cuDNN disabled, greedy decoding, thinking disabled, a 64-token output cap,
and batch size one. The input token IDs are frozen before generation; the
largest input is 2,475 tokens and no input is truncated. Generation uses
PyTorch 2.7.1+cu118 and Transformers 5.10.2. An initial incompatible CUDA
runtime produced no predictions; the repaired environment uses byte-identical
prepared inputs and a separately recorded runtime configuration.

All 2,400 predictions completed without a terminal error and were sealed
before reference answers were loaded. Independent replay reconstructs every
input token sequence from the original messages, decodes every output token
sequence, and reproduces all native substring exact-match scores. It checks
artifacts and scoring; it does not rerun GPU generation or the original
source extraction.

CMC, fact BM25, and stale closure score 96.50/76.50, 96.50/20.50, and
32.00/16.50 percent on single-/multi-hop questions, respectively.
Against fact BM25, CMC gains 230 multi-hop answers and loses six, with
164 ties. Mean Qwen reader-input lengths are 201.39, 1,794.00, and
197.38 tokens for the three arms. This experiment requires no new source
extraction. Its tokenizer-specific counts are reported separately from HY4.

\paragraph{Llama reader on the same extracted facts.}
A separate full-population run uses the frozen CMC and fact-BM25 evidence
and question messages with Llama-3.1-8B-Instruct, checkpoint
\texttt{0e9e39f249a16976918f6564b8830bc894c89659}, its own chat
template, and a 64-token output cap. All 1,600 predictions were freshly
generated, sealed, natively scored, and independently replayed. CMC answers
363/400 single-hop and 220/400 multi-hop questions; fact BM25 answers
368/400 and 46/400. On multi-hop questions CMC gains 184 paired answers
and loses ten. This is a reader replacement on fixed extracted evidence.

\subsection{Same-backbone comparison with SH-Conflict/CAR}
\label{app:car-comparison}

\begin{table}[ht]
\centering
\small
\caption{Same-backbone FactConsolidation comparison: 400 SH and 400 MH questions.
Both systems receive the original histories and use HY4-preview.
The adapted released SH-Conflict/CAR methods retain their query-dependent
extraction and stage prompts; CMC uses source-only extraction and one reader
call. Frontends and output caps differ.}
\label{tab:car-comparison}
\begin{tabular}{lrr}
\toprule
Method & SH & MH \\
\midrule
SH-Conflict / CAR (adapted) & 98.00 & 60.75 \\
CMC & 97.25 & 78.25 \\
\bottomrule
\end{tabular}
\end{table}

We run the released SH-Conflict method for single-hop questions and CAR for
multi-hop questions on all 800 original FactConsolidation questions and
histories. The source code is pinned to revision
\texttt{7d319f460b0ee0945d7de05d06c34681dceca46a}.
The adapter preserves the released prompts, fact segmentation,
first-serial deduplication, BM25 top-10 retrieval (\texttt{rank-bm25}
0.2.2 defaults), and the loop that skips unavailable hop dependencies and
returns the last valid answer. It replaces the model transport with
HY4-preview at temperature zero, \texttt{no\_think}, JSON-object output,
and a 4,096-token output cap per structured call. Telemetry and gold-based
evaluation are removed from inference. The final answer comes from the
released loop, with no additional reader call.

The population, source projection, code, and model settings were frozen
before generation. All 800 predictions completed before labels were loaded;
none ended in a transport or parsing error. A separate check replays the
released algorithm with saved responses, verifies its requests and traces,
and recomputes the released substring exact-match scores. This checks
execution and scoring without rerunning the model.

SH-Conflict answers 392/400 single-hop questions correctly and CAR answers
243/400 multi-hop questions correctly. CMC answers 389 and 313, respectively.
The paired CMC gains/losses are 1/4 on single-hop and 79/9 on multi-hop;
395 and 312 outcomes tie. Thus CMC improves multi-hop accuracy by 17.50
points while single-hop accuracy is 0.75 points lower.

SH-Conflict uses 400 calls totaling 182,099 input and 25,494 output tokens.
CAR uses 1,301 calls totaling 662,831 input and 79,255 output tokens.
Every call has provider usage recorded, with one transport attempt per
call. These counts include the query-dependent decomposition and extraction
steps. CMC instead prepares a source-only extraction shared across questions
and makes one reader call per question; its frontend cost is reported in
Appendix~\ref{app:llm-protocols}. Different frontends, prompts, and output
caps make this an end-to-end accuracy comparison. The shared-extraction
controls isolate evidence selection separately.

\section{BM25 at the CMC Evidence Budget}
\label{app:matched-budget}

This control uses the same frozen direct-LLM extraction, current facts,
unresolved records, HY4 reader, and 800 questions as the shared-extraction
experiment. For each question, its evidence allowance equals the cl100k
token count of CMC's rendered evidence, including the prefix and record
numbers. This tokenizer defines the packing budget; actual HY4 input usage
is measured separately.

We first reserve every unresolved record. We then visit the first 100 facts
in the original BM25 ranking, adding a fact only when the complete rendered
context fits the allowance and the 60,000-character cap. Each trial restores
source order and re-encodes the whole string. Nonfitting facts are skipped;
records are never cut and the candidate list is never extended. The mean
evidence length is 103.89 cl100k tokens, against CMC's 106.93. All 800
questions retain nonempty evidence.

\begin{table}[ht]
\centering
\small
\caption{All length groups in the budget comparison. Each cell contains
100 questions scored with the released substring exact-match metric.}
\label{tab:matched-budget-lengths}
\setlength{\tabcolsep}{4.5pt}
\begin{tabular}{lrrrrrrrr}
\toprule
& \multicolumn{4}{c}{Single-hop} & \multicolumn{4}{c}{Multi-hop} \\
\cmidrule(lr){2-5}\cmidrule(lr){6-9}
Evidence & 6K & 32K & 64K & 262K & 6K & 32K & 64K & 262K \\
\midrule
CMC & 98 & 98 & 98 & 95 & 90 & 77 & 85 & 61 \\
Fact BM25 & 98 & 98 & 98 & 96 & 33 & 19 & 20 & 11 \\
BM25, matched cap & 97 & 97 & 98 & 96 & 9 & 6 & 9 & 5 \\
\bottomrule
\end{tabular}
\end{table}

On the 400 multi-hop questions, CMC is correct on 286 items where matched
BM25 is incorrect; the reverse occurs on two items. On those 286 items the
matched pool never contains the full CMC-selected fact set and contains the
scored CMC answer string on 19 items. Their mean actual reader
inputs, pooled over all 800 questions, are 184.64 and 182.22 tokens.
The control uses 727 new calls and 73 identical same-question CMC responses.
All outputs are completed before scoring, and both complete 800-question
controls are replayed. New reader usage is 147,246 tokens; the reused
reused responses account for another 8,185 tokens. Shared extraction and the
computation of CMC's per-question budget are additional costs. This is an
evidence-length control, not a comparison of total computation.

\section{MQuAKE-MemStream: Construction and Validity}
\label{app:mquake-stream-protocol}

\paragraph{Source and transformation.}
MQuAKE-MemStream derives from the released MQuAKE-Remastered counterfactual
CF-3k cases \citep{zhong2025mquakeremastered}. We retain the released
question wording and answer labels.
For every case, we append each \texttt{single\_hops} cloze followed by its
released answer to a pre-edit block, and each \texttt{new\_single\_hops}
cloze followed by its released answer to a post-edit block. Each
\texttt{requested\_rewrite} contributes its subject--relation template
filled with the old target to the first block and the new target to the
second. Consequently, the constructed memory explicitly contains the
intermediate and final answer-bearing support facts. The requested
rewrite may repeat a single-hop assertion; we keep both observations.
Every method receives a source bank derived from these histories. This
controlled test measures version handling and evidence selection with
supplied support facts, not discovery from unrestricted documents.

\paragraph{History and question construction.}
Cases are taken in released file order and pooled in groups of 100.
Pre-edit assertions are shuffled within a block and post-edit assertions
within a second block, using seed 20260907; the pre-edit block precedes
the post-edit block. This ordering is imposed by our adaptation: MQuAKE
does not provide a chronological agent-memory stream. The position in our
constructed stream identifies the newer observation.
Each case contributes its first released composite question and one
released post-edit single-hop question sampled with the same random
generator. The dataset uses the first 800 cases, all two-hop. Before the
consistency check, this gives 800 candidate questions per class. The final
dataset has 798 multi-hop and 800 single-hop questions. Answer aliases are
stored separately for scoring and do not select graph paths.
All 5,560 assertions in the 800-case Qwen streams are parsed by the
relation grammar, so these results do not test extraction from open-form
language. The supplementary dataset includes the source cases, constructed
streams, questions, answers, and exclusions. The accompanying code
reconstructs the dataset and evaluates the released predictions.

\paragraph{Source interpretation and evidence views.}
The existing relation-grammar parser processes the constructed assertions.
CMC resolves the latest parsed edge for each subject--relation key and
traverses from explicit query subjects up to five hops. The latest-state
control returns all current assertions. Stale closure retains CMC's
selected keys and substitutes their most recent earlier different object
where one exists. The raw-history control retains every version before
applying the reader's window policy. All views preserve original serial
numbers. Raw history exceeds the 4,096-token Qwen reader input and is
trimmed from its oldest end.

\paragraph{Pool-consistency audit.}
For every case, we compare each released post-edit single-hop relation with
the final value of its subject--relation key after pooling. Two of the 800
candidate multi-hop cases have a hop overwritten by another case. We
exclude their multi-hop questions from the named dataset. This rule uses
released support facts and no model predictions. We apply it to the stored
per-question predictions for every arm. Before this exclusion, multi-hop
accuracy was 96.38\% for CMC, 53.25\% for latest state, 13.63\% for stale
closure, and 60.38\% for raw history. The filtered results are in
Table~\ref{tab:mquake-stream}. Extending the same construction to
all 3,000 CF-3k cases alters at least one released post-edit hop in 431
cases. We therefore do not use that larger run as confirmatory evidence.
The supplementary code checks these exclusions and the larger-pool
sensitivity count.

\paragraph{Interpretation.}
The constructed dataset and its supplied support facts differ from the official
edit-count protocol, which tests a model after specified knowledge edits;
our task tests reading a supplied memory. The scores are reported
separately and must not be compared directly with official MQuAKE scores.
Pooling allows assertions from different cases to interact. Scoring retains
the released case answers, and a case answer is not used to repair the pooled
history or its current graph.

\section{Traversal before Value Refresh}
\label{app:order-control}

\paragraph{Constructed prior graph.}
The control uses the same frozen LLM extraction as CMC. For each
subject--relation key, we choose the most recent earlier assertion whose
object differs from the current object. If none exists, we keep the current
assertion. This constructs a prior value for each key; it is not a single
historical snapshot. Both graphs have the same keys and explicit query
anchors. We traverse the constructed graph to depth five, then replace
each selected key's value with its current assertion. We render these
current source records together with the same unresolved records and
60{,}000-character budget used by CMC. Thus all returned parsed facts
are current, while the selected keys can differ.

\paragraph{Population and reader.}
All 800 questions and requests were fixed before new generation. The
source-only key sets differ on 659 questions: 290 single-hop and 369
multi-hop. They agree on 141 questions. All contexts fit the evidence
budget. Exact full-request matches on the same question permit 141
fixed reuses; the remaining 659 requests use HY4-preview with temperature
zero, thinking disabled, and a 64-token output cap. The existing 800 CMC
requests were reconstructed exactly. All 800 outputs were collected before
scoring, and every question remains in the denominator.
The HY4 alias is unpinned, so the recorded reader settings do not establish
an immutable model checkpoint across execution dates.

\begin{table}[t]
\centering
\small
\caption{Complete order control with frozen extraction and HY4-preview. Accuracy is substring exact match (\%). Key-change strata are computed from source graphs before scoring.}
\label{tab:order-control}
\begin{tabular}{lrrrr}
\toprule
Population & $n$ & Prior + refresh & CMC & Difference \\
\midrule
Single-hop & 400 & 97.25 & 97.25 & 0.00 \\
Multi-hop & 400 & 21.50 & 78.25 & 56.75 \\
MH: keys unchanged & 31 & 100.00 & 100.00 & 0.00 \\
MH: keys changed & 369 & 14.91 & 76.42 & 61.52 \\
\bottomrule
\end{tabular}
\end{table}

Both views score 97.25\% on single-hop questions. On multi-hop questions,
CMC scores 78.25\% and the control 21.50\%, with 235 CMC-only correct
answers and eight control-only correct answers. Among the 369 multi-hop
questions whose keys change, the scores are 76.42\% and 14.91\%.
Both answer all 31 unchanged-key multi-hop questions correctly.
Mean reader-input lengths are 184.64 and 194.56 tokens, respectively.
The 659 new calls use 138,246 input and 3,943 output tokens; all complete
without transport retries. No new source-extraction calls are made.

This comparison tests selection under a constructed version perturbation
on the public development grid. It complements stale closure, which holds
the selected keys fixed and changes their values. A separate implementation
reconstructs both sets of 800 requests and replays all native scores.

\section{Traversal Depth with Fixed Extraction}
\label{app:hop-depth}

We vary the traversal cap over $h\in\{1,2,3,5\}$ while keeping the
LLM extraction, current-state construction, anchors, source-record renderer,
60{,}000-character cap, and HY4 reader fixed. Each setting uses all 800
FactConsolidation questions. Five hops reuses the original CMC outputs;
the other three settings contribute 2,400 new reader calls.

Single-hop accuracy changes little: 96.75\% at one hop and 97.25\% at
the other tested caps. Multi-hop accuracy rises from 5.50\% at one hop to
54.00\%, 71.75\%, and 78.25\% at two, three, and five hops.
The corresponding mean reader inputs are 115.29, 138.42, 158.41, and
184.64 tokens. Expanding beyond the named entity therefore recovers
useful evidence at a modest input-length increase on this grid.

We reconstructed the 3,200 contexts and requests from the fixed extraction
and source histories, checked saved predictions, and reproduced the scores.
The five-hop results reuse the same completed reader requests as the main
experiment.

\section{Four-Competency Evaluation Protocol}
\label{app:mab-full-protocol}

\paragraph{Population and aggregation.}
The full-suite experiment uses the released MemoryAgentBench arXiv v3 main-table
population: 1,000 accurate-retrieval questions, 700 test-time-learning
questions, 171 long-range-understanding questions, and 200 fact-consolidation
questions. These span 14 source datasets and 130 histories. The SF population
contains the 262k single- and multi-hop histories; the shorter SF histories
remain in the separate mechanism analysis. Overall performance averages the
four competencies. Within TTL, we first average the five classification
datasets and then average classification and recommendation. We retain every
scheduled question, including failed generations and invalid judgments.

\paragraph{Dataset provenance.}
MemoryAgentBench reconstructs SH-Doc QA and MH-Doc QA from RULER
\citep{hu2026memoryagentbench,hsieh2024ruler}. It introduces EventQA and
FactConsolidation; the latter uses MQuAKE counterfactual pairs
\citep{zhong2023mquake}. Its LongMemEval (S*) task reconstructs
LongMemEval conversations \citep{wu2024longmemeval}.
The five classification tasks draw on BANKING77
\citep{casanueva2020banking77}, CLINC150 \citep{larson2019clinc150},
the multi-domain NLU dataset \citep{liu2019nlu}, and TREC coarse and
fine question classification \citep{li2002trec}. Recommendation uses
ReDial \citep{li2018redial}. The two long-range tasks use
$\infty$Bench En.Sum \citep{zhang2024infinitebench} and DetectiveQA
\citep{xu2024detectiveqa}. These are the source datasets; the question
populations and scoring follow the released MemoryAgentBench adaptation.

\paragraph{Source interpretation and CMC records.}
The four interfaces are a recent-history suffix, raw-block BM25, dense/raw
rank fusion, and the CMC interfaces described below.
Source interpretation receives original history text. Public task instructions
and questions enter reading; reference answers and evaluation metadata enter
only scoring. Labeled training demonstrations are legitimate source history
on TTL tasks. Accurate retrieval uses an original-window interface: a frozen
overlapping-window partition of the original source, retrieved with BM25.
The card frontend used for DetectiveQA and global summaries
partitions histories into blocks of at most 12,000 characters and parses
batches of up to five blocks into at most 40 attributed cards, using a
6,000-token output cap. One source-only prompt serves those batches.
Card quotations are checked against their attributed source blocks.

Of 1,961 card-frontend batches, 120 pass completely, 1,027 partially, and
814 yield no cards, mainly because their JSON cannot be parsed. The parser
retains 20,640 quote-checked cards. Summary reads fall back to marked
original excerpts when cards are absent. These counts measure validation,
not semantic completeness. The main accurate-retrieval row uses window BM25.

All interfaces share a 60,000-character evidence budget and the same HY4
reader contract within each task. Dense and card-guided lookup use
equal-weight reciprocal ranks with constant 60; dense rankings use
normalized BGE-M3 CLS embeddings with an 8,192-token cap
\citep{chen-etal-2024-m3}.
\paragraph{Original-window read.}
Accurate retrieval and recommendation use the same window selector.
The index stores source spans of at most 2,048 characters, preferring
nearby line endings and retaining 256 characters of overlap.
Window size and overlap are fixed for AR and recommendation. For each question,
BM25 ($k_1=1.2$, $b=0.75$) retrieves 64 candidate windows.
The renderer selects up to 24 windows under the 60,000-character budget,
merges overlapping spans, and restores source order. The reader receives
the selected original text with source coordinates. LongMemEval contexts
are first reconstructed into dated speaker sessions; sessions after the
question's Current Date are omitted. The main CMC row uses BM25 window
order for every AR column and for recommendation Recall@5 (17.8). These
five columns use the same window selector, chosen during development.
MCC, Summary, and DetectiveQA in
Table~\ref{tab:mab-hy4-main} are the original same-reader campaign: MCC
86.4/85.8/86.2/86.2 and DetectiveQA use card/raw block lookup, Summary
uses coverage notes, and DetQA scores use the registered JSON-field
decoder (76.1/77.5/78.9/67.6). Matched raw BM25 uses
12,000-character blocks instead of these identity records.

Source indexing is shared across questions; candidate scoring is
query-specific.
The SF CMC and history-suffix requests reuse 400 completed predictions only
after exact request-identity checks. Source parsing and question answering
costs are recorded separately. Global-summary coverage records
representation and clipping; it does not guarantee preservation of
every source fact.

The full campaign records 37.62 million tokens for shared source processing,
27.99 million for new and reused CMC reader requests, and 2.38 million for
CMC evaluation judgments. These totals exclude unavailable failed-attempt
usage, offline embedding, and inherited SF extraction.

In FactConsolidation, CMC uses 252 and 282 mean reader-input tokens for
single- and multi-hop questions, against about 15{,}000 for the history
suffix. Reader-input counts exclude source preparation, local retrieval,
and evaluation judgments.

The 60,000-character cap omits some original blocks in every broad-suite
original-block read. It also truncates all 200 SF history-suffix inputs, but
none of the 200 CMC SF inputs. All 100 CMC summary reads clip generated
notes and include original-text fallbacks. Recorded output-token caps do
not establish that answers were truncated; finish reasons were not recorded.

\paragraph{Scoring and comparison scope.}
Non-judge metrics use the released scoring functions. LME and summarization
use the released judgment prompts with HY4 as judge. This is a matched HY4
comparison with local generation contracts, rather than a reproduction of
the original generation and judge models. EventQA's primary metric is the
documented substring exact match. A correction registered during source
processing, before reader generation, selects this field from the native scorer; the initially
configured event-recall field remains a diagnostic. Native DetectiveQA exact match is retained as a diagnostic of that format
interaction. The matched DetQA column uses the registered JSON-field decoder
described below. Answer content and scores never trigger retries. Published
results are distinguished by their reported model and evaluation protocol.

\paragraph{What the evaluation does and does not cover.}
The four-competency grid tests the released MemoryAgentBench tasks under a
shared reader. Other memory evaluations probe different conditions: LoCoMo
tests very long conversations \citep{maharana-etal-2024-evaluating}, ConvoMem
varies conversational scale and temporal changes \citep{pakhomov2025convomem},
MemoryArena links tasks across sessions \citep{he2026memoryarena}, and
LongMemEval-V2 tests reuse of web-agent experience
\citep{wu2026longmemevalv2}. Mem2ActBench asks agents to apply remembered
information to tool actions \citep{shen-etal-2026-mem2actbench}.
Temporal question answering itself spans several task formulations
\citep{piryani2026tqasurvey}. Our result on ordered, single-valued
FactConsolidation assertions should not be read as a score on these other
benchmarks.

The fixed CMC read contract also differs from adaptive memory control.
BudgetMem learns which memory budget tier to use
\citep{zhang2026budgetmem}; MemFlow routes by intent
\citep{chen2026memflow}; RouterMem conditions further execution on evidence
sufficiency \citep{lin2026routermem}; and MemCon learns a policy over
retrieval and consolidation \citep{jiang2026memcon}.
MemAgent and InfMem instead optimize sequential reading and memory updates
for very long inputs \citep{yu2026memagent,wang2026infmem}.
These policies can issue further reads or change their stored state; the
CMC mechanism comparison holds source extraction and the answer call fixed.

Other systems also make different assumptions about updates. StateFuse keeps
conflicting observations visible to a resolver \citep{volkov2026statefuse},
while TrustMem verifies that consolidation preserves and supports stored
information \citep{yang2026trustmem}. Infini Memory maintains topic
documents \citep{ji2026infinimemory}, and MemCompiler compiles memory for
embodied state-conditioned execution \citep{ding2026memcompiler}.
CMC's last-write-wins rule is specific to the benchmark's ordered,
single-valued relations; it does not resolve simultaneous values or
multi-agent conflicts.

\paragraph{DetectiveQA output-format diagnostic.}
\label{app:mab-detective-format-protocol}
DetectiveQA requests a JSON object with an answer field. Before reader
outputs were generated, we registered a decoder that passes the complete
response to Python's \texttt{json.loads} and scores its string-valued
\texttt{answer} field when present; otherwise it scores the original
response. It does not strip code fences or select text using reference
answers. Native exact match is zero in all four arms, whereas decoded exact
match is 76.06, 77.46, 78.87, and 67.61 for the history suffix, BM25,
dense+BM25, and CMC, respectively. The main-table DetQA cells use these
decoded scores. All 71 items per arm remain in the denominator, including
failed generations at zero.

\paragraph{Completed summary-judge diagnostics.}
All 1,200 summary-judge requests returned. On the 100 summary questions,
12 history-suffix and 73 CMC outputs have invalid judgments. Each affected
item has an unparseable precision judgment at the 4,096-token cap; invalid
items retain zero in the primary table. CMC summaries exceed the requested
1,200-word limit on 76 items, versus 16--23 in the other arms, and 20 CMC
reader outputs reach their generation cap. These observations suggest an
output-format interaction, but do not establish what another judging contract
would score. All 1,200 LongMemEval judgments are valid, so this issue does
not explain the accurate-retrieval regressions.

\section{Runtime Audit}
\label{app:runtime-audit}

We measured the complete FactConsolidation population: eight histories and
800 questions, with the direct-LLM extraction frozen before timing. A local
Python 3.11 process used \texttt{perf\_counter\_ns} on a Linux host with an
Intel Xeon Platinum 8255C CPU and 20 available cores. Each history was
replayed 20 times; the timing includes
normalization, latest-value replacement, and residual-record collection.
The measured replay excludes source loading, LLM extraction, and the BM25
index used by a control. We then initialized the shared-extraction view and
checked all 800 rendered contexts for each of CMC and fact BM25 against
their frozen answer-call inputs. Each question was read five times per arm
in shuffled order. CMC timing includes explicit anchor matching, five-hop
closure, and source rendering. Fact BM25 timing includes ranking of current
facts and source rendering, without CMC-only graph work in the shared
dispatcher. Both exclude answer generation; fact BM25's one-time index
construction is also excluded. The reported percentiles pool the two
histories at each length. The supplementary code includes the timing script.

\begin{table}[H]
\centering
\caption{CPU time for state replay and evidence selection on frozen
extraction. Replay has 40 runs, and each read arm has 1{,}000 runs per
history length. Read columns give median / 95th percentile in milliseconds.}
\label{tab:runtime-cpu}
\begin{tabular}{lrrr}
\toprule
History length & CMC replay (s) & CMC read (ms) & Fact BM25 read (ms) \\
\midrule
6K   & 0.004 & 0.35 / 0.42 & 0.47 / 0.59 \\
32K  & 0.023 & 1.61 / 1.96 & 1.79 / 2.15 \\
64K  & 0.047 & 3.32 / 3.90 & 3.35 / 4.24 \\
262K & 0.204 & 13.78 / 144.34 & 14.33 / 17.96 \\
\bottomrule
\end{tabular}
\end{table}

At 262K, the two read medians are close, but CMC has a higher 95th
percentile. The current CMC implementation rebuilds graph adjacency for
each read; these measurements do not establish a CPU latency advantage.

We also summarize request durations already recorded by the HY4-preview
reader runs. They include transport and provider-side response time, while
excluding local rate-limit queueing. These historical calls were not
interleaved across arms, and prompt-cache use differed, so their durations
are descriptive service observations rather than a controlled latency
comparison. The CMC arm used the exact inputs validated above. Offline LLM
extraction was a separate set of experiments with 827 calls, 919{,}509
input tokens, and 419{,}813 output tokens
(Appendix~\ref{sec:llm-frontend}). This work includes alignment and
revision, so it is not a cold-build latency measurement. The present
numbers are not a full end-to-end wall-clock cost.

\begin{table}[H]
\centering
\caption{Observed answer-request duration in recorded HY4-preview calls,
800 completed requests per arm. Prompt-cache fraction is the share of prompt
tokens recorded as cached.}
\label{tab:runtime-reader}
\begin{tabular}{lrrr}
\toprule
Evidence arm & Median (s) & p95 (s) & Cached prompt fraction \\
\midrule
CMC          & 1.29 & 1.95 & 0.00 \\
Fact BM25    & 1.42 & 3.15 & 0.003 \\
Latest state & 2.46 & 4.88 & 0.557 \\
\bottomrule
\end{tabular}
\end{table}

\end{document}